\documentclass{article}

 \usepackage{arxiv}

\usepackage[utf8]{inputenc} 
\usepackage[T1]{fontenc}    
\usepackage{microtype}
\usepackage{graphicx}
\usepackage{color}
\usepackage{xcolor}
\usepackage{hyperref}
\hypersetup{
    colorlinks,
    linkcolor={red!50!black},
    citecolor={blue!50!black},
    urlcolor={blue!80!black}
}
\usepackage{url}            
\usepackage{booktabs}       
\usepackage{amsfonts}       
\usepackage{nicefrac}       
\usepackage{microtype}      
\usepackage{enumitem}
\usepackage[round]{natbib}
\usepackage{doi}
\usepackage{amsmath,bbm,dsfont}
\usepackage{cleveref}       

\newcommand{\bbox}{\begin{tcolorbox}[colback=red!5!white,colframe=red!75!black,boxsep=0mm]}
\newcommand{\ebox}{\end{tcolorbox}}

\renewcommand{\paragraph}[1]{\textbf{#1}\,\,}
\newcommand{\squeeze}{\looseness=-1}

\newcommand{\red}[1]{{\leavevmode\color{red}{#1}}}
\newcommand{\blue}[1]{{\leavevmode\color{blue}{#1}}}
\newcommand{\green}[1]{{\leavevmode\color[RGB]{0,128,0}{#1}}}

\newcommand{\cyan}[1]{{\leavevmode\color{cyan}{#1}}}
\newcommand{\darkblue}[1]{{\leavevmode\color[RGB]{0,0,115}{#1}}}

\newcommand\todo[1]{{\red{TODO: {#1}}}}

\newcommand\tocitec[1]{\red{[CITE: {#1}]}}

\newcommand\extended[1]{} 
\newcommand\priority[1]{} 

\newcommand{\niradd}[1]{\green{#1}}

\newcommand{\hf}[1]{{\cyan{[Haifeng: #1]}}}

\newcommand\expect[2]{\mathbbm{E}_{#1}{\left[ {#2} \right]}}

\newcommand{\one}[1]{\mathds{1}{\{{#1}\}}}

\DeclareMathOperator*{\argmax}{argmax}

\newcommand{\yhat}{{\hat{y}}}

\newcommand{\vhat}{{\hat{v}}}

\newcommand{\dist}{{D}}

\newcommand{\policy}{{\pi}}
\newcommand{\choice}{{c}}
\newcommand{\reward}{{r}}

\newcommand{\welf}{{\mathtt{welfare}}}

\title{Welfare as a Guiding Principle for Machine Learning ---
From Compass, to Lens, to Roadmap}

\author{   Nir Rosenfeld   \\
	Faculty of Computer Science \\
	Technion -- Israel Institute of Technology\\
	\texttt{nirr@cs.technion.ac.il}  \\
	\And
	 Haifeng Xu \\
	Department of Computer Science\\
    University of Chicago \\ 
	\texttt{haifengxu@uchicago.edu}  \\
}

\renewcommand{\shorttitle}{Machine Learning Should Maximize Welfare}

\hypersetup{
pdftitle={Machine Learning Should Maximize Welfare},
pdfauthor={Nir Rosenfeld, Haifeng Xu},
pdfkeywords={welfare, machine learning},
}

\begin{document}

\maketitle

\begin{abstract}

Decades of research in machine learning have given us powerful tools for making accurate predictions.
But when used in social settings and on human inputs,
better accuracy does not immediately translate to better social outcomes.
To effectively promote social well-being through machine learning,
this position paper advocates for the wide adoption of \emph{social welfare} as a guiding principle.
The field of welfare economics asks:
how should we allocate limited resources to self-interested agents in a way that maximizes social benefit?
We argue that this perspective applies to
many modern applications of machine learning in social contexts.
As such, we propose that welfare serves
as an additional core criterion in the design, study, and use of learning algorithms,
complementing the conventional pillars of
optimization, generalization, and expressivity,
and as a compass guiding both theory and practice.



\end{abstract}

\section{Introduction} \label{sec:intro}



As the influence of machine learning on our lives continues to grow,
there is high hope that this will prove to be for the better.
Given its unmatched ability of to make accurate predictions,
we have, in principle, much reason for optimism.
Accurate predictions have the potential
to improve the choices we make for ourselves,
from the mundane (e.g., what to buy or where to eat)
to the highly consequential (e.g., what to study, how to invest, where to live, or what career to pursue).
This helps explain why machine learning has become the backbone of most
recommendation systems,
media services,
online marketplaces,
and social platforms.
Better prediction can also support decisions made about us, such as
which medical treatment to apply,
when to offer financial aid,
or who to hire.
Learned models are therefore making their way,
slowly but surely,
into more traditional social domains
such as health care, 
education,
finance,
transportation,
law,
and even government.
Given how learning increasingly shapes
the ways we communicate, express ourselves, and are productive---%
its influence and integration are only likely to grow.
This calls for careful consideration of the role we envision for machine learning in society---as it is forming now, and as we would like it to be.

The problem is that machine learning is designed to map inputs to outputs:
it has no intrinsic capacity to account for broader context,
and lacks a principled framework for reasoning about its social aspects and implications.
Learning already builds on multiple foundations, including
probability and statistics for generalization,
numerical analysis for optimization,
and approximation theory 
for expressivity.
Similarly, 
it requires
an additional toolkit for addressing the social challenges
arising from its growing integration within society.
We argue that welfare economics---through its central notion of \emph{social welfare}---%
can serve as an effective framework for this purpose, both conceptually and rigorously.
\squeeze

\textbf{Our position is that social welfare should be the \emph{lens}
through which we design, analyze, and evaluate learning algorithms in social contexts.
It should serve as a \emph{compass} providing concrete, well-defined goals,
and as a \emph{roadmap} for how to develop and apply welfare-aware learning algorithms.}
\emph{Welfare} should be established
as an additional core learning principle,
complementing and enhancing
the canonical pillars of \emph{optimization}, \emph{generalization}, and \emph{expressivity}.
In contrast to existing principles which
are extensively studied and widely understood, the economic foundations of learning remain far less developed---%
despite their significant implications and clear importance.
Learning algorithms should be designed to explicitly maximize the well-being of users by accounting for economic context,
and enabling policymakers to implement diverse social policies.
Economic pitfalls and failure modes should be discussed, scrutinized, and mitigated---%
as routinely and systematically as overfitting, local minima, and model misspecification are addressed;
in social settings, they should carry equal importance
and receive comparable attention in both research and practice.
\squeeze

We believe that the notion of social welfare can, and should,
be accessible and relevant to 
the broader machine learning community,
and in particular to any researcher or practitioner
whose work bears social impact.
We envision a roadmap of adopting the welfare principle including 
shaping the community's operational and methodological norms; 
guiding the design, evaluation, and deployment of learning systems in practice;
informing the reviewing process and publication criteria;
becoming an integral part of machine learning course curricula;
enabling an effective interface with policymakers;
and, ultimately, 
taking center stage in academic discussions on the future of machine learning in society.
Throughout the paper we attempt to make these ideas  more precise,
better grounded, and concretely contextualized.
\squeeze

We illustrate the gaps between current learning approaches
and what welfare requires---and enables---%
through a series of learning tasks of increasing economic complexity,
beginning with accuracy maximization (with awareness to welfare)
and gradually transitioning to explicit welfare maximization
(via prediction). 
These examples serve demonstrate how welfare can serve as a lens for guiding the design and analysis of learning algorithms and their interface with policy.
We discuss the potential merits of this perspective, 
the technical and conceptual challenges it presents,
the steps needed to support its successful adoption,
and the limitations that are likely to remain.


 \subsection{Machine learning: an economic perspective}

If we accept that machine learning has the capacity to be socially beneficial, 
then ideally, we would like it to improve outcomes for everyone.
But economic theory casts doubt as to the feasibility of such an aspiration.
A basic economic intuition is that when people benefit from something,
it often becomes a scarce resource---over which they then contend.
Our key point, which we will argue throughout,
is that \emph{learning in social contexts is inherently, and inevitably,
about how  some  form of limited resource is allocated}.
We illustrate this through various examples,
from clearly limited resources such as loans and jobs,
to more nuanced cases like recommended and generated content,
and even users.

As a result,
learning algorithms increasingly act as determinants of allocation,
effectively making
decisions about who gets what---and who doesn't. 
As conventional learning framework pursues ever-improving accuracy,
we are possibly missing out (or even deteriorating) the ability to make effective and beneficial allocations of the real to-be-allocated resources. We posit that  many instances in which the use of machine learning has led to undesired social outcomes, even if unintentional, can be explained by accuracy itself being limited
(for instance, see \citep{wang2024counterfactual}). 
\squeeze

When demand exceeds available resources,
the inevitable reality is that
not everyone can get a piece of the cake.
This naturally gives rise to competition---between the system and its users,
amongst users themselves,
and between competing service providers.
\extended{or all of the above.}
Conventional learning tools,
which treat human inputs as any other `static' abstract input,
fail to account for the impact of \emph{human agency}---%
through individual knowledge and behavior, 
competition, and cooperation---%
on learning and its outcomes.
As a result, the sense of control and foresight we have come to expect of learning systems to provide can break down, becoming  unreliable,
and risking misalignment with reality.

Our goal is to promote the study of learning algorithms that are aware of the economic context in which they operate, and are explicitly designed to promote favorable social outcomes---by accounting for
resource scarcity and human agency, and how they relate.
Towards this, our main thesis is that learning frameworks should be designed to innately support a notion of \emph{social welfare}.
In economics, welfare quantifies the overall benefit to agents in an economic system,
and questions regarding the estimation, maximization, and distribution of welfare are central to the analysis of any such system.
It also intrinsically supports the idea that agents act autonomously to promote their own interests, and provides a concrete means for policymakers to balance trade-offs, steer outcomes, and implement diverse policies.
\squeeze

\squeeze

\subsection{Limitations of current practice: motivating example}

It is tempting to hope that more data, more compute, and more sophisticated learning algorithms will pave the way to a better tomorrow.
But merely improving predictions, whether to match or go beyond human capabilities,
still presents inherent limitations.
Consider the following example:


\textit{%
A school receives funding to provide students with targeted aid intended to help them pass an exit exam required for graduation.
To decide whom to select for the program,
the school collects historical data on students who participated in earlier programs and trains a model to predict when aid is likely to be effective.
It then uses this model to determine which new students to admit.
}

On its face, this approach appears reasonable---%
but matters are more subtle:
the school may wish to maximize the impact of aid,
but doing so by 
maximizing accuracy falls short in several key respects:%
\footnote{Other issues are apparent in the example, e.g., distribution shift or causal effects---but these are orthogonal to our argument.
\squeeze
}
\begin{itemize}[leftmargin=1em,topsep=0em,itemsep=0.2em]

\item 
Maximizing accuracy corresponds to maximizing the number of successes.
This does not consider \emph{who} receives aid, nor how its benefits are distributed across students.

\item 
Students who are selected may choose not to participate.
As a result, the program may operate below capacity,
leaving resources unused, and reducing overall effectiveness.

\item 
Conversely, when space is limited, students are likely to compete for acceptance.
This can bias selection towards those better able to \emph{appear} qualified---or to game the system.

\item
Selection is based on whether aid will be effective.
It does not account for whether, or how much, \emph{students themselves} benefit from passing the exam.

\end{itemize}




This example illustrates that while accuracy
can contribute to social benefit,
it does not fully capture it.
As we argue, this gap has \emph{structure} that manifests
broadly whenever learning is applied in social contexts.
Social welfare captures this structure through a modular framework comprising several core components:
\emph{resources}, through their availability and scarcity;
\emph{allocations} and their distribution across the population;
\emph{user agency}---through knowledge, incentives, and actions;
\emph{decision-making} and its downstream effects;
and \emph{social policy}, in both range of supported policies
and the means available to policymakers.
We advocate for the study and advancement of welfare in learning
along these axes,
both independently and in their interaction with other core learning principles.

\subsection{What has been studied, and what is still missing}


Although welfare has been considered to some extent in several subfields of machine learning,
its use has so far been limited in scope and context,
focusing only on particular aspects,
and in service of specific agendas.
We next discuss these lines of research from a welfare viewpoint:
what they capture, what they overlook,
and why they may fail to address even basic welfare issues.





\paragraph{Fairness.}
Social good in machine learning is commonly associated with
the fairness literature \citep{dwork2012fairness}.
Fairness constraints enable making benefits equal,
but cannot express or promote economic efficiency---i.e., the overall benefit generated \citep{heidari2018fairness}.
Whereas fairness captures a particular notion of equity (by limiting variance),
welfare provides a framework for explicitly balancing equity and efficiency (by maximizing expected outcomes).
Fairness therefore yields a restricted space of social policies,
narrowing policymakers' role to deciding what to equalize.
A second major gap in this literature is its limited acknowledgment of human agency.
For example, predictions for school admissions may satisfy demographic parity,
but unless members of the minority group \emph{choose} to apply---%
believing it is in their best interest---%
such guarantees may fail in practice \citep{horowitz2024classification}.
Works in this domain that discuss welfare mostly portray it 
as an alternative means for evaluating fair outcomes
\citep{hu2020fair,kasy2021fairness,cousins2021axiomatic}.
This acknowledges that \emph{something} is in limited supply---%
typically positive predictions---%
but does not capture the broader economic context of resource scarcity.
Although some recent works explicitly model allocation through prediction
\citep{rolf2020balancing,shirali24allocation,perdomo2024the},
human agency remains largely absent.

\paragraph{Strategic and performative learning.}
The notion of user agency
is central to fields like
strategic classification
\citep{bruckner2012static,hardt2016strategic}
and performative prediction \citep{perdomo2020performative},
which model user behavior explicitly or through its indirect effects on dynamics or equilibrium.
Nonetheless, work in these field has largely revolved around accuracy maximization, with user behavior 
typically portrayed as an obstacle---e.g., by modeling users as ``gaming'' the system.
In contrast, welfare views accuracy as a means for promoting user value, rather than an end in itself.
Some works do consider social aspects,
such as social burden \citep{milli2019social}
or transparency \citep{ghalme2021strategic},
\extended{\todo{add scmp, scai, recdiv}}
and others aim to support cooperation \citep{levanon2022generalized}
or improvement \citep{kleinberg2020classifiers,shavit2020causal},
but these typically treat user outcomes as a consequence of accuracy efforts.
The focus on accuracy also implies that correct predictions are the (implicit) scarce resource.
While this simplifies user responses,
it does not align with the field's main motivating examples---%
such as loans, admissions, and hiring \citep{liu2022strategic}---%
where relevant resources (e.g., funds, slots, and positions) impose constraints that can induce
complex dependencies and
externalities
(e.g., admitting one student reduces opportunities for others).
A final shortcoming is that despite its potential for realistic applications,
the field remains mostly theoretical.
\squeeze

\paragraph{LLMs and Alignment. } Following the
success of large language models (LLMs),
there has been an explosive interest in their \emph{alignment} \citep{ouyang2022training,liutrustworthy}, i.e., tuning them to better match human preferences. Although alignment is directly motivated by the goal of improving welfare across the user population, research in this space has not explicitly integrated welfare as a core design principle. Instead, most current research casts alignment as a reward maximization problem,
mostly via singular, scalar reward functions  \citep{ouyang2022training,rafailov2023direct},
through recently also using pluralistic objectives \citep{sorensenposition,feng2024modular}.
Additional complications arising from limited model capacity
and user behavior---for example that users \emph{choose} which LLM to work with---have been largely overlooked.
Only recently, efforts have been made to move beyond the 
the ``harmlessness--helpfulness'' paradigm of LLM alignment and explicitly account for user agency \citep{naseem2026llm}
or competition between service providers \citep{wei2026market}.

\section{Compass: Welfare economics and what it can teach us} \label{sec:welfecon}

Welfare economics is the subfield of economics that is concerned with the 
characterization, evaluation, and maximization of social welfare in economies and societies.
The main principles of welfare economics date back to Adam Smith \citep{smith1759moral,smith1776wealth},
but its formal foundations were laid out only a century later by
notable neoclassic economists such as
  \cite{edgeworth1881psychics},
\cite{marshall1890principles},
\cite{pareto1906manual},
and \cite{pigou1920welfare}.
Modern welfare economists include influential figures such as
 \cite{hicks1939value},
 \cite{arrow1951social}, \cite{sen1970collective},
and  \cite{stiglitz2012inequality}---all of which received the Nobel Prize for their contributions to this field.
Given its rich history,
we believe that machine learning has much to gain 
from this well-established discipline.

\subsection{Welfare economics: crash course}
\paragraph{The distribution of wealth.}
The main question welfare economics asks is:
how should wealth be distributed across individuals in the economy?
Under the working hypothesis that resources (and therefore wealth) are limited,
the main object of interest in welfare economics is the \emph{Pareto front}---%
the set of all possible economic states in which no individual can be made better off without making things worse for another \citep[see, e.g.,][]{johansson1991introduction}.
In terms of welfare, there are two main considerations for policymakers:
\squeeze
\begin{enumerate}[leftmargin=1.7em,topsep=0em,itemsep=0.3em]
\item \textbf{Efficiency:}
How do we reach a Pareto state?

\item \textbf{Equity:}
Of all Pareto states, which are preferable? 
\end{enumerate}
\emph{Efficiency} requires an ability to optimize economic outcomes 
to a point where social benefit is maximal (in the Pareto sense).
Markets are a classic example of how the actions of many self-interested agents can combine to produce  efficient outcomes \citep{arrow1954existence}.
However, there are typically many states 
that are maximally beneficial,
but that differ in how benefits are distributed across individuals;
in regard to this, markets are mostly silent.
As such, \emph{equity} makes a statement about the relative preference ordering over all possible states, and considers means for steering towards preferable ones.
A canonical example is income distribution:
all governments likely seek higher overall income (efficiency),
but may disagree about whether high inequality should be permitted or suppressed
(equity).

\priority{%
\todo{add pareto curve graphic? general econ and/or acc-vs-welfare?}
}

\paragraph{Social welfare functions.}
In welfare economics,
the primary tool for defining and promoting efficiency and equity 
is the \emph{social welfare function},
which ranks or scores all possible economic states.
Our focus will be on the common choice of cardinal (i.e., real-valued) welfare functions that take the form of an expectation over weighted individual utilities
(see discussion and alternatives in Appx.~\ref{appx:SWFs}): %
\squeeze
\begin{equation}
\label{eq:welfare}
\welf(\policy;w) = \expect{(x,y)\sim \dist}{w(x) u(x,y;\policy)}
\end{equation}
where $\policy$ is a \emph{policy}
determining 
which resources are allocated to whom,
and $u(x,y;\policy)$ is the utility of user $x$ with label $y$ under $\policy$.
Note $w(x)$ is a weight function that the planner \emph{chooses}:
this defines
the desired direction for overall improvement (i.e., efficiency)
by balancing the importances of individual outcomes (i.e., equity). %
Throughout we will consider
policies $\policy=\policy_h$ guided by a predictive model $h$.
\squeeze


\paragraph{The social planner.}
Equity is inherently a subjective notion that requires making value judgements.
Social welfare functions make it possible to formally express these by setting appropriate weights.
In welfare economics, weights are designated by a \emph{social planner}---either a real or fictitious entity that represents societal preferences.
A social planner can set weights to aid low income individuals;
implement affirmative action towards some social group;
or ensure that all individuals obtain some minimal level of utility.
Concrete examples include using weights to prioritize certain subgroups \citep{bjorkegren2022machinelearningpoliciesvalue} or balance different objectives \citep{rolf2020balancing}.
The simplest weighing scheme is of course using uniform weights, i.e., $w(x) = 1$ for any $x$.
But note even this makes a statement, which is that individuals should be weighted by their utility;
this is known as `utilitarian welfare'.
Hence, from the perspective of welfare economics, any objective that optimizes a (non-weighted) average---such as accuracy in machine learning---is in effect making a statement about how value should be distributed.

\paragraph{Human agency.}
Welfare economics makes explicit the idea that individuals have \emph{agency}.
Intuitively, this states that individuals
(i) \emph{want} things,
(ii) \emph{know} things,
and (iii) \emph{do} things---to get what they want, using what they know.
These notions are formally accounted for by modeling
utility functions or preferences (want),
private information and beliefs (know),
and decision-making, e.g., rational or behavioral (do).
Any policy that aims to advance welfare must take these into account.
Often this requires the planner to make additional efforts,
such as to elicit preferences,
create incentives for truthful reporting,
or infer how users will respond to different policy choices.
These are challenging, but give the planner power:
if incentives can be aligned, then it becomes possible
to harness the willingness of individuals to invest effort for improving outcomes for all; consider public goods and services, crowdfunding platforms, open-source software, and collaborative knowledge bases.
\squeeze

\section{Lens: A welfare perspective of  machine learning in social contexts}
We believe that machine learning has much to gain from adopting a welfare perspective:
when inputs represent humans, it becomes 
possible to promote overall social benefit (efficiency),
and imperative to consider its distribution across the population (equity).
Welfare provides a principled framework for reasoning about and advancing these goals,
while exposing missing or implicit assumptions that hinder their achievement.
To demonstrate, we begin with the most basic learning objective of supervised binary prediction, and show how welfare can serve
as a guiding principle.
We then gradually move towards objectives that
support increasingly complex economic considerations.

\subsection{Accuracy as a social welfare function}
Consider the conventional setting of supervised classification:
Let $\dist$ be an unknown joint distribution over pairs $(x,y)$ describing the features and labels of individuals (or generally `users') in a population.
Given a training set of labeled samples $S = \{(x_i,y_i)\}_{i=1}^m \sim \dist^m$,
the goal is to find a function $h$ from a chosen class $H$
whose predictions $\hat{y}=h(x)$ 
maximize the expected accuracy on future examples:
\squeeze
\begin{equation}
\label{eq:accuracy_objective}
\argmax\nolimits_{h \in H} \expect{(x,y) \sim \dist}{\one{y = h(x)}}
\end{equation}
Since the objective in Eq.~\eqref{eq:accuracy_objective} is an expectation over the population,
it can be interpreted as prescribing a particular social welfare function---%
a special case of Eq.~\eqref{eq:welfare}---%
in which:
(i) users derive utility from correctness of individual predictions,
(ii) all users share the same utility function,
$u(x,y;\policy_h) = \one{y = h(x)}$,
(iii) possible outcomes include correct and incorrect predictions,
hence (iv) predictions inform policy directly as $\policy_h(x)=h(x)$,
and finally (v) weights are uniform, $w(x) = 1$ for all $x$.

Despite the apparent `neutrality',  
the above reveals that accuracy maximization in fact makes an (implicit)
statement about social preferences.
This entails a particular notion of equity---%
one that derives operationally from the predictive task at hand,
rather than from a planned or designated social agenda.
\squeeze

\paragraph{Resources.}
Since welfare concerns the allocation of resources,
a first question is: what resources are scarce and contested?
This, in turn, depends on what drives user utility.
Across diverse applications, a common source is individual accuracy---%
from content recommendation and asset pricing
to medical diagnosis and career planning.
In these cases, \emph{accuracy itself is the limited resource},
and the choice of classifier becomes a decision on which users will be `allocated'  correct predictions, and which will not.
\squeeze

Notice that the `amount' of potentially
available resources depends on the chosen model class $H$.
This interpretation sheds light on the role of traditional learning principles:
\emph{optimization} concerns maximally utilizing this potential by finding the optimal $h \in H$ over the training set;
\emph{generalization} ensures that resource availability
guarantees transfer to
test time;
and \emph{expressivity} serves to increase the available resource pool by 
enabling richer function classes.
Note all the above essentially concern efficiency,
namely how \emph{many} users are allocated resources---but not \emph{which}
users are allocated.

\paragraph{Allocation.}
Given that accuracy is scarce, the next question is:
who \emph{should} be allocated accurate predictions?
One answer is users who most need it, or who benefit mostly.
As a simple example, assume users differ in their potential gain from correctness.
This entails the following welfare objective:
\squeeze
\begin{equation}
\label{eq:order0_objective}
\argmax\nolimits_{h \in H} \expect{(v,x,y) \sim \dist}{v \cdotp \one{y = h(x)}}
\end{equation}
where $v$ is the value of accuracy of user $x$ with label $y$, now sampled jointly from $\dist$.
This transforms Eq.~\eqref{eq:accuracy_objective} into an explicit utilitarian welfare objective in which utility derives from individual prediction correctness:
$u(x,y;h) = v$ if $h(x)=y$, and zero if not.
The solution to Eq.~\eqref{eq:order0_objective} is no longer to maximize the amount of allocations,
but rather, to allocate in a way which generates maximal social gain.
\squeeze

\paragraph{Agency.}
It is tempting to consider Eq.~\eqref{eq:order0_objective} as a simple weighted accuracy objective,
or more generally as an instance of cost-sensitive classification \citep[e.g.,][]{elkan2001foundations}.
The crux is that this neglects user \emph{agency}:
since values are private to users (`know'),
and since users seek accurate predictions (`want'),
they are prone to overreporting values (`do') as this can
tilt allocations in their favor.
This introduces uncertainty into example weights, but
due to economic factors
rather than statistical.
Thus, unless the learning process accounts for agency,
outcomes can be very different
from those depicted by the objective.
\squeeze

\subsection{Welfare as an explicit learning objective} \label{sec:lens-explicit}

Although accurate predictions are beneficial,
user utility often depends on downstream decisions.
As we discuss next,
this introduces challenges even when learning explicitly targets social welfare.


\paragraph{Resources and allocation.}
Consider a system that makes decisions about individuals
through predictions
in order to maximize social welfare.
For a user $x$ with label $y$, denote by $a$ the action the system takes,
and assume the decision policy depends on predictions as
$a = \policy(h(x)) = \policy_h(x)$,
for example by thresholding on predicted scores.
These are generally known as \emph{prediction policies} \citep{kleinberg2015prediction}.

\extended{\todo{say something about causality or DFL, or dodge?}}

When decisions are modeled explicitly, resource constraints arise naturally.
These can express, for example,
a limited number of available jobs (via cardinality constraints),
a total sum of funds that a bank can lend (knapsack constraints),
regulation on the amount of financial risk an insurer can take (bounded expected risk),
or a limit on the number of posts a social platform can block while enabling free speech (lower-bounded rates).
Given a set $A$ of feasible allocations,
the objective can be:
\begin{equation}
\label{eq:order1_objective}
\argmax\nolimits_{h \in H} 
\expect{\dist}{u(\policy_h(x),y)}
\quad \text{s.t.} \quad
\policy_h(\dist) \in A
\end{equation}
where
$u(a,y)$ encodes utility to user $x$ with label $y$ from
action $a$,
and
$\policy_h(\dist)$ is the distribution of actions over the population.  
To focus our discussion,
for now we will abstract away the uncertainty in $u$.
\squeeze


\paragraph{Agency.}
Clearly, machine learning has many tools for coping with constraints,
and approaches such as top-$k$ classification \citep[e.g.][]{lapin2015top,petersen2022differentiable} or learning to rank \citep[e.g.][]{cao2007learning}
may seem adequate for Eq.~\eqref{eq:order1_objective}.
Unfortunately, such methods are not designed to account for user agency
and its implications under limited resources.
As a simple example, 
assume
users benefit from positive predictions as 
$u(x,y;\policy_h)=v \cdot \one{h(x)=1}$. 
Because utility stems from predictions,
users will act---in response to $h$---to improve their own predictive outcomes.
But since resources are limited,
users must \emph{compete with one another} to secure them.
Consider for example admissions under a limited quota:
applicants cannot simply meet a fixed acceptance threshold,
but must instead outperform others---i.e., the bar adjusts with competition.
\extended{This suggests that systems have less control over outcomes than typically assumed
and depicted by conventional learning objectives.}%
The challenge is that competition introduces complex dependencies across users' behavior, known
as \emph{externalities}.
Maximizing welfare therefore requires learning models that effectively coordinate the collective behavior of users---
akin to the classic role of markets.

\extended{As for \emph{how} users behave,
while the literature on strategic learning focuses primarily on feature manipulation (i.e., changing $x$ to some other $x'$ at a cost),
in reality users clearly engage in a much broader range of behaviors.
For example, a similar yet distinct behavior is to misreport $x$;
this carries no direct costs, but could risk a penalty if detected.
A different type of behavior is 
choosing weather to at all participate---%
known as \emph{self-selection} \citep{roy1951some}.
While extensively studied in econometrics,
it is fairly new to machine learning
\citep{zhang2021classification,cherapanamjeri2023makes,horowitz2024classification,saig2026evolutionary}.
Users can also  misreport $y$ \citep{meir2012algorithms},
withhold certain features \citep{krishnaswamy2021classification},
or influence others \citep{eilat2023strategic,gois2025performative};
each behavior type carries its own implications.}

\paragraph{Information.}
The actions of agents are inherently guided by their knowledge and beliefs.
Some of this information is private---such as valuations (as above), true feature values,
additional unobserved attributes, internal mental states, and personal hopes, aspirations, and expectations.
These give users an informational advantage over the system.
Unfortunately, information asymmetry 
is the root of common
market failures,
including moral hazard (hidden actions),
adverse selection (hidden features),
and costly signaling (hidden features, observable action); more on these below.
Maximizing welfare therefore requires that learning objectives effectively accommodate information gaps.

\subsection{Welfare in the hands of a social planner} 

Although welfare can serve as an ideal to aspire to,
realistically many systems will pursue different goals.
Clear examples are online social, media, or commercial platforms who seek profit.
\extended{And while revenue in such platforms builds on user satisfaction,
it need not fully align with welfare.}%
This calls for learning objectives that enable a social planner
to balance between system goals and social benefit.

\paragraph{Resources and allocation.}
When a platform's gains depend on users' behavior,
\emph{users themselves} often become the contested resource---%
through their time, attention, or engagement. 
Allocation is therefore no longer fully controlled by the system,
but also depends on users' subsequent choices.
The concern is that platforms may enlarge their share by means other than increasing
user value---for example, by driving excessive engagement or overconsumption \citep{mujica2022addiction}.
This relates to the economic notion of \emph{phishing equilibria},
coined by Nobel laureates Akerlof and Shiller \citep{akerlof2015phishing},
in which firms profit by exploiting consumers' cognitive biases and psychological vulnerabilities.

\paragraph{Agency.}
In commercial settings, outcomes for users are more often mediated by predictions
rather than determined by them directly.
For example, predictions can shape a set of recommended alternatives,
but utility ultimately depends on how users choose from the set.
Denoting the system's goal (e.g., revenue) by $\reward$,
and denoting (model-mediated) user choices by $\choice_h$,
a possible learning objective is:
\squeeze
\begin{equation}
\label{eq:order2_objective}
\argmax\nolimits_{h \in H} 
\expect{\dist}{\reward(\choice_h(x,y); \policy_h)}
\end{equation}
User choice behavior can range from selecting items or content within a platform 
to choosing the platform itself. 
This highlights that learning often takes place in a competitive environment
in which service providers and platforms compete over user market share.
While recently of growing interest to the learning community \citep{ben2020content,jagadeesan2023learning}, 
this aspect is typically abstracted away.

\extended{with learning objectives depicting a world where providers are static or in which only one platform exists.}


\paragraph{Information.}
Platforms, too, may keep certain information private---a key example is the learned model itself.
While transparency is often viewed as normatively desirable,
a welfare perspective can help determine when it is also socially beneficial in a utilitarian sense---and when it is not \citep{ghalme2021strategic}.
One concern is that platforms' control over information disclosure can be used
to shape user beliefs and steer behavior towards their own goals,
echoing ideas from \emph{information design} \citep{bergemann2019information}. 
However, when exercised responsibly and under appropriate regulation,
such control can also be used to balance
the platform's proprietary interests with the well-being of its users.
\squeeze

\paragraph{Social planner.}
Once platforms use learning to promote their own objectives,
promoting welfare requires a third party---namely a social planner---to align incentives
towards societal preferences.
Whether endogenous to the platform (e.g., management) or exogenous (e.g., government),
planners must account not only for the platform's actions and constraints,
but also for user agency, as well as that of other parties
such as content creators, sellers, service providers, etc.
The role of machine learning in this context should be to \emph{enable} planners
to implement their policy---whatever it might be.
Thus, learning objectives should be flexible and support a wide spectrum of possible policies,
spanning a wide and efficient Pareto front from which a planner can choose a desired operating point.

\extended{
\blue{
Technically, objectives such as Eq.~\eqref{eq:order2_objective}
can support welfare for example by adding it as a constraint (e.g., lower bounding $\welf(h)$) or as regularization (e.g., adding the term $\lambda \cdot \welf(h)$).}

\blue{can view the learner as "agent" to which welfare applies;
  welfare describes wellbeing of all - including eg sellers}
}


\section{Roadmap: Welfare as a core guiding principle} 
Machine learning is a discipline born of many fields,
whose diverse tools are essential for coping with the complexities
of real-world learning tasks.
Over time, key organizing principles have emerged---such as 
optimization, generalization, and expressivity---that  enable reasoning 
about its core aspects.
They provide a language for communicating ideas and insights,
as well as for identifying pitfalls and scrutinizing limitations
in the process of developing solutions.
For example, describing a learning algorithm as
"hard to optimize", "prone to overfitting", or "not expressive enough"
makes clear what is problematic, as well as what should be done about it.
Such principles also help us make informed decisions,
for instance assessing the implications of convexity (or its absence);
the cost-benefit of acquiring additional data;
or the trade-offs in increasing the depth of a neural network.

What remains missing is a guiding principle that similarly addresses
the fundamental aspects that arise when learning algorithms are deployed in social settings.
Our main thesis is that social welfare can serve as a useful framework for this purpose,
alongside and complementing existing principles.
We next elaborate on how a welfare perspective can enable this, and what it brings to the table.

\paragraph{Principal axes.}
Working within a welfare framework requires specifying and understanding
each of its main components:
resources, allocation, agency (through incentives, knowledge, and actions),
decisions, and social policy (via the social planner and welfare function)---%
both individually and in their interactions.
This enables asking guiding questions, such as: what is scarce? why are resources contested? and what can users do about it?
Answers to these questions shed light on the role that learning
plays in shaping social outcomes,
and inform the design of appropriate learning objectives.
It also prompts questions that uncover missing and latent aspects,
for instance: what is the implied social welfare function?
what is implicitly assumed about user behavior?
or what information is private to each party?
While it is both valid and often useful to restrict focus to a subset of
these elements, a welfare perspective requires being clear on what
is targeted, and what is assumed away.
By comparison, an empirical paper can have merit even without providing explicit
sample complexity bounds---yet generalization is still an issue that is expected to be clearly addressed and discussed.

\extended{
\blue{%
for every ml project, begin by asking: what is scarce? who allcoates and how? what do people want, know, and do? how do these align with system incentives? what polices *can* be implemented, and how?
}

\todo{table comparing welfare to opt, gen, exp along dimensions?}
}

\paragraph{Failure modes.}
Overfitting, bad local minima, and model misspecification
are common ways in which learning can fail to achieve its goals.
These failure modes are well-understood and central to both research and applied work.
In contrast, deploying machine learning in social settings entails
new economic risks that are much less understood,
but which a welfare perspective can help mitigate.

Some economic failure modes stem from agency:
Moral hazard arises when users take on excessive risk and shift costs onto the system (e.g., credit card overuse);
adverse selection arises when one side exploits
its informational advantage over others
(e.g., in car sales or medical insurance);
and costly signaling arises when decision-making requires
agents to take costly actions to convey crucial private information
(e.g., to identify qualified applicants for job hiring).
Other failure modes pertain to the (im)balance of power---%
such as monopolies, oligopolies, and consumer lock-in;
to collective action---%
such as free riding, tragedy of the commons, and coordination failure;
or to policy and incentives---%
such as Goodhart's law, Campbell's law, the Lucas critique, and the cobra effect.

We conjecture that many of these can materialize once learning systems exert influence over social outcomes.
The central questions are when and which failures arise, 
what facilitates or exacerbates them,
and how learning algorithms can be 
better designed to account for and manage them.
We see value in studying these questions theoretically and practically
in both stylized and realistic settings.
\squeeze


\extended{\todo{%
- interconnectedness and interaction between welfare and opt, gen, exp:
  - example: l2 regularization
  - example: SGD implicit bias
}}

\paragraph{Tools of the trade.}
It is hard to imagine designing machine learning algorithms
without tools from probability 
or linear algebra.
Tackling additional economic challenges requires an adequate toolkit;
not surprisingly, economics has such tools to offer.
Modeling learning as a game between the system, its users, and possibly others actors
can help capture the nature of their relations.
For example, a Stackelberg game enables modelling users as responding independently to a learned model (as in strategic classification \citep{levanon2021strategic,horowitz2023causal});
a Stackelberg-Nash game \citep{liu1998stackelberg}
models interdependent user responses
(e.g., through a learning-induced market \citep{sommer2025learning} or externalities \citep{hossain2025strategic});
and a Nash-Stackelberg game \citep{sherali1984multiple,bernheim1986common} can model learners competing over users
(e.g., content creators competing for user attention \citep{yao2023bad}).
\squeeze

Mechanisms help capture the structure of interactions,
such as in terms of information and actions,
and  explain both the sources of incentive misalignment and ways to correct them.
For example,
\emph{contracts} facilitate cooperation by aligning incentives
(e.g., for delegating learning tasks \citep{saig2023delegated});
\emph{auctions} support efficient resource allocation under private valuations \citep{myerson1981optimal};
\emph{markets} enable effective coordination among decentralized agents
(e.g., in markets for 
accuracy \citep{ben2019regression,einav2025a},
organic content \citep{jagadeesan2023competition,yao2023bad,yao2024user},
model-generated content \citep{keinan2026strategic},
or general preferences \citep{wei2026market});
and \emph{matching markets} 
coordinate bilateral preferences
to yield efficient outcomes based on mutual choices 
\citep{jagadeesan2023learning,dai2024incentive}.  

Many mechanisms are either designed to explicitly promote welfare
or to support its organic emergence 
\citep[e.g.,][]{arrow1954existence,shapley1971assignment}.
A central question is whether this holds also when allocations are mediated by predictions
\citep{nahum2024decongestion}.
Given the growing concerns about how prediction-based recommendations and generative content
may drive polarization, echo chambers, informational barriers, economic inequity, and unhealthy usage patterns, the answer is likely negative.
We posit that these phenomena can be interpreted as certain types of market failure.
A welfare perspective can help elucidate the role of learning in these processes
and guide the design of effective monitoring and regulatory interventions.

\paragraph{Synergy.}
Despite the economic nature of welfare challenges,
we believe that solutions can emerge from within machine learning itself,
through its native tools, methods, and practices.
This draws inspiration from fields that have successfully adapted 
predictive machinery to solve complex downstream tasks, such as
decision-focused learning \citep{mandi2024decision},
algorithms with predictions \citep{mitzenmacher2022algorithms},
prediction-powered statistical inference \citep{angelopoulos2023prediction},
and causal inference \citep{wager2018estimation}.
For example, equity concerns can arise from learning objectives focused solely on expected outcomes.
Yet underspecified objectives are a recurring issue in
robustness \citep{d2022underspecification},
explainability \citep[see][]{rudin2019stop},
and fairness \citep{rodolfa2020case,coston2021characterizing,black2022model}.
Luckily, work on model multiplicity 
\citep{breiman2001statistical,marx2020predictive,hsu2022rashomon}
offers insights into potential remedies.
Since user behavior can often be cast as 
\emph{decision-dependent distribution shift} \citep{drusvyatskiy2023stochastic},
tools from 
distributional robustness \citep{chen2020distributionally}
and OOD
generalization \citep{buhlmann2020invariance}
can help tame resulting outcomes.
The growing interest in expressing economic mechanisms via differentiable surrogates \citep{bichler2025differentiable}
enables smooth integration into neural networks as implicit layers \citep{bai2019deep}
or via differentiable solvers \citep{agrawal2019differentiable,berthet2020learning}.
Social planners can use
methods for end-to-end learning of Pareto fronts \citep{navon2021learning}---%
or even regularization \citep{levanon2021strategic}---%
to navigate 
the space of implementable policies.
\squeeze

\extended{while navigating this space can sometimes be as simple as using regularization \citep{levanon2021strategic}.}

Beyond existing tools,
we anticipate welfare will motivate new integrative challenges for researchers and practitioners across many subfields of machine learning;
some examples can be found in \citep{jordan2026a},
which advances a `collectivist' economic perspective of machine learning.
But machine learning does not only offer methodological `solutions'---%
it offers a whole package of norms, conventions, and standard practices;
consider for example how competitions and benchmarks drive advances in the field,
or the constructive interplay between attacks and defenses in adversarial learning.
With some creativity and ingenuity efforts,
these could also be harnessed to advance welfare-aware machine learning.
\squeeze

\section{Alternative views} \label{sec:alt_views}
While it is easy to agree that learning systems should be designed to promote social good, disagreement about \emph{how} to achieve this is only natural.
We propose to adopt the perspective of welfare economics
but there certainly exist complementary approaches and
viewpoints; see also Appendix~\ref{appx:beyond}.

\extended{We highlight possible alternative viewpoints below, and discuss additional prospects and challenges in Appendix~\ref{appx:beyond}.}

\paragraph{Give accuracy time.}
One perspective is that if we wait long enough for sufficient data to accumulate and existing methods to improve, then machine learning will organically overcome the challenges we discussed.
One example to draw on is how despite many advances in optimization,
the basic idea of gradient descent (and its variations) still drives many modern tools.
Another is how large language models have demonstrated that simply predicting the next word with sufficient accuracy and on enough data 
gives rise to emergent phenomena far beyond this basic task.
Our position is that resource scarcity is an inherent problem of any social system, whether technology-driven or not.%
\extended{Our point here would be that since even 50 thousand years of cultural evolution have not `solved' the problem of scarce resource allocation, 
it is unlikely to just go away.}
We believe that scarcity and agency should be addressed explicitly---%
but of course we may be proven wrong.
\squeeze

\paragraph{Divide and conquer.}
Even if  machine learning as a standalone solution does not suffice,
one could argue that an economic approach can be applied disjointly,
rather than be integrated. 
Hence, learning and policy can be advanced independently and combined only later.
This is reasonable, and independent efforts and application will likely be required regardless. 
But there is increasing evidence that this will not suffice;
in fact, the field of fair machine learning rose in response to the clear need for embedding social considerations within the learning objective itself.
Advances in the study of fairness in learning have also shown that fairness constraints alone cannot guarantee equity,
such as when learning effects accumulate over time \citep[e.g.,][]{liu2018delayed}.
We take these to suggest that the novelty in the interface between learning and economics requires a wholistic approach
specialized for this intersection.%
\extended{that operates on a deep, joint understanding of how both fields can intersect.}
\squeeze

\paragraph{Welfare without welfare.}
Welfare economics is not the only approach for reasoning about and facilitating welfare,
nor is it free from issues and limitations in itself.
Criticism includes
its subjective nature;
the need to measure and compare utility across individuals; 
the emphasis on cardinal rather than ordinal utilities;%
\extended{(which are more plausible but harder to work with);}
the reliance on assumptions of rational behavior; 
the practical challenges of accounting for complex externalities and market failures;
the need for a centralized social planner entity;
and challenges in policy evaluation.
Other schools of thought in economics offer alternatives:
For example, Sen's \emph{capabilities approach}
\citep{sen1999commodities}
focuses on ensuring people are capable of achieving what they seek,
rather than the value of what they obtain.
Within machine learning, there been have calls for alternative approaches as well,
such as to `democratize' the issue of alignment using social choice theory
\citep{conitzer2024position,ge2024axioms,fish2024generative}.
We view these as complementary to ours,
and believe there is merit advancing our collective well-being through machine learning
simultaneously along several fronts.

% ... 





 
\section{Discussion} \label{sec:discussion}

Economics studies the tensions that arise from having limited means but unlimited wants.
Our main thesis is that while the appetite for what machine learning can give us only grows,
we lack the tools to integrate the structure inherent in social settings into the design of learning algorithms.
Welfare economics offers a framework for designing systems that enable a social planner to express and promote diverse social policies subject to 
resource scarcity and user agency.
Our goal has been  to identify and illustrate how this perspective can, with appropriate adaptations, guide both research and practice in machine learning for social settings.
Scarcity in learning-driven environments can be far more elusive than in traditional economic settings with tangible goods;
nonetheless, it is present, and its implications are central to effectively leveraging machine learning for increased social benefit.
Though the path forward may be long and challenging,
we believe welfare should serve as the compass, roadmap, and lens
for making informed choices about the role we envision for machine learning in society.
\squeeze

\bibliographystyle{plainnat}
\bibliography{refs.bib}

\begin{thebibliography}{100}
\providecommand{\natexlab}[1]{#1}
\providecommand{\url}[1]{\texttt{#1}}
\expandafter\ifx\csname urlstyle\endcsname\relax
  \providecommand{\doi}[1]{doi: #1}\else
  \providecommand{\doi}{doi: \begingroup \urlstyle{rm}\Url}\fi

\bibitem[Adam{~}Smith(1759)]{smith1759moral}
Adam{~}Smith.
\newblock \emph{The Theory of Moral Sentiments}.
\newblock A. Millar, London, 1759.

\bibitem[Adam{~}Smith(1776)]{smith1776wealth}
Adam{~}Smith.
\newblock \emph{An Inquiry into the Nature and Causes of the Wealth of
  Nations}.
\newblock W. Strahan and T. Cadell, London, 1776.

\bibitem[Adler(2019)]{adler2019measuring}
Matthew~D Adler.
\newblock \emph{Measuring social welfare: An introduction}.
\newblock Oxford University Press, USA, 2019.

\bibitem[Agrawal et~al.(2019)Agrawal, Amos, Barratt, Boyd, Diamond, and
  Kolter]{agrawal2019differentiable}
Akshay Agrawal, Brandon Amos, Shane Barratt, Stephen Boyd, Steven Diamond, and
  J~Zico Kolter.
\newblock Differentiable convex optimization layers.
\newblock \emph{Advances in neural information processing systems}, 32, 2019.

\bibitem[Akerlof(1978)]{akerlof1978market}
George~A. Akerlof.
\newblock The market for “lemons”: Quality uncertainty and the market
  mechanism.
\newblock In \emph{Uncertainty in economics}, pages 235--251. Elsevier, 1978.

\bibitem[Akerlof and Shiller(2015)]{akerlof2015phishing}
George~A Akerlof and Robert~J Shiller.
\newblock Phishing for phools: The economics of manipulation and deception.
\newblock 2015.

\bibitem[Angelopoulos et~al.(2023)Angelopoulos, Bates, Fannjiang, Jordan, and
  Zrnic]{angelopoulos2023prediction}
Anastasios~N Angelopoulos, Stephen Bates, Clara Fannjiang, Michael~I Jordan,
  and Tijana Zrnic.
\newblock Prediction-powered inference.
\newblock \emph{Science}, 382\penalty0 (6671):\penalty0 669--674, 2023.

\bibitem[Arrow(1951)]{arrow1951social}
Kenneth~J. Arrow.
\newblock \emph{Social Choice and Individual Values}.
\newblock Wiley, New York, 1951.

\bibitem[Arrow and Debreu(1954)]{arrow1954existence}
Kenneth~J. Arrow and Gerard Debreu.
\newblock Existence of an equilibrium for a competitive economy.
\newblock \emph{Econometrica: Journal of the Econometric Society}, pages
  265--290, 1954.

\bibitem[Bai et~al.(2019)Bai, Kolter, and Koltun]{bai2019deep}
Shaojie Bai, J~Zico Kolter, and Vladlen Koltun.
\newblock Deep equilibrium models.
\newblock \emph{Advances in neural information processing systems}, 32, 2019.

\bibitem[Ben-Porat and Tennenholtz(2019)]{ben2019regression}
Omer Ben-Porat and Moshe Tennenholtz.
\newblock Regression equilibrium.
\newblock In \emph{Proceedings of the 2019 ACM Conference on Economics and
  Computation}, pages 173--191, 2019.

\bibitem[Ben-Porat et~al.(2020)Ben-Porat, Rosenberg, and
  Tennenholtz]{ben2020content}
Omer Ben-Porat, Itay Rosenberg, and Moshe Tennenholtz.
\newblock Content provider dynamics and coordination in recommendation
  ecosystems.
\newblock \emph{Advances in Neural Information Processing Systems},
  33:\penalty0 18931--18941, 2020.

\bibitem[Bergemann and Morris(2019)]{bergemann2019information}
Dirk Bergemann and Stephen Morris.
\newblock Information design: A unified perspective.
\newblock \emph{Journal of Economic Literature}, 57\penalty0 (1):\penalty0
  44--95, 2019.

\bibitem[Bernheim and Whinston(1986)]{bernheim1986common}
B~Douglas Bernheim and Michael~D Whinston.
\newblock Common agency.
\newblock \emph{Econometrica: Journal of the Econometric Society}, pages
  923--942, 1986.

\bibitem[Berthet et~al.(2020)Berthet, Blondel, Teboul, Cuturi, Vert, and
  Bach]{berthet2020learning}
Quentin Berthet, Mathieu Blondel, Olivier Teboul, Marco Cuturi, Jean-Philippe
  Vert, and Francis Bach.
\newblock Learning with differentiable pertubed optimizers.
\newblock \emph{Advances in neural information processing systems},
  33:\penalty0 9508--9519, 2020.

\bibitem[Bichler and Parkes(2025)]{bichler2025differentiable}
Martin Bichler and David~C Parkes.
\newblock Differentiable economics: Strategic behavior, mechanisms, and machine
  learning.
\newblock \emph{Communications of the ACM}, 68\penalty0 (8):\penalty0 80--88,
  2025.

\bibitem[Bj{\"o}rkegren et~al.(2020)Bj{\"o}rkegren, Blumenstock, and
  Knight]{bjorkegren2020manipulation}
Daniel Bj{\"o}rkegren, Joshua~E Blumenstock, and Samsun Knight.
\newblock Manipulation-proof machine learning.
\newblock \emph{arXiv preprint arXiv:2004.03865}, 2020.

\bibitem[Björkegren et~al.(2022)Björkegren, Blumenstock, and
  Knight]{bjorkegren2022machinelearningpoliciesvalue}
Daniel Björkegren, Joshua~E. Blumenstock, and Samsun Knight.
\newblock ({M}achine) learning what policies value.
\newblock \emph{arXiv preprint arXiv:2206.00727}, 2022.

\bibitem[Black et~al.(2022)Black, Raghavan, and Barocas]{black2022model}
Emily Black, Manish Raghavan, and Solon Barocas.
\newblock Model multiplicity: Opportunities, concerns, and solutions.
\newblock In \emph{Proceedings of the 2022 ACM Conference on Fairness,
  Accountability, and Transparency}, pages 850--863, 2022.

\bibitem[Breiman(2001)]{breiman2001statistical}
Leo Breiman.
\newblock Statistical modeling: The two cultures (with comments and a rejoinder
  by the author).
\newblock \emph{Statistical science}, 16\penalty0 (3):\penalty0 199--231, 2001.

\bibitem[Br{\"u}ckner et~al.(2012)Br{\"u}ckner, Kanzow, and
  Scheffer]{bruckner2012static}
Michael Br{\"u}ckner, Christian Kanzow, and Tobias Scheffer.
\newblock Static prediction games for adversarial learning problems.
\newblock \emph{The Journal of Machine Learning Research}, 13\penalty0
  (1):\penalty0 2617--2654, 2012.

\bibitem[B{\"u}hlmann(2020)]{buhlmann2020invariance}
Peter B{\"u}hlmann.
\newblock Invariance, causality and robustness.
\newblock \emph{Statistical Science}, 35\penalty0 (3):\penalty0 404--426, 2020.

\bibitem[Cao et~al.(2007)Cao, Qin, Liu, Tsai, and Li]{cao2007learning}
Zhe Cao, Tao Qin, Tie-Yan Liu, Ming-Feng Tsai, and Hang Li.
\newblock Learning to rank: from pairwise approach to listwise approach.
\newblock In \emph{Proceedings of the 24th international conference on Machine
  learning}, pages 129--136, 2007.

\bibitem[Chen et~al.(2020)Chen, Paschalidis, et~al.]{chen2020distributionally}
Ruidi Chen, Ioannis~Ch Paschalidis, et~al.
\newblock Distributionally robust learning.
\newblock \emph{Foundations and Trends{\textregistered} in Optimization},
  4\penalty0 (1-2):\penalty0 1--243, 2020.

\bibitem[Conitzer et~al.(2024)Conitzer, Freedman, Heitzig, Holliday, Jacobs,
  Lambert, Moss{\'e}, Pacuit, Russell, Schoelkopf,
  et~al.]{conitzer2024position}
Vincent Conitzer, Rachel Freedman, Jobst Heitzig, Wesley~H Holliday, Bob~M
  Jacobs, Nathan Lambert, Milan Moss{\'e}, Eric Pacuit, Stuart Russell, Hailey
  Schoelkopf, et~al.
\newblock Position: Social choice should guide {AI} alignment in dealing with
  diverse human feedback.
\newblock In \emph{Forty-first International Conference on Machine Learning},
  2024.

\bibitem[Coston et~al.(2021)Coston, Rambachan, and
  Chouldechova]{coston2021characterizing}
Amanda Coston, Ashesh Rambachan, and Alexandra Chouldechova.
\newblock Characterizing fairness over the set of good models under selective
  labels.
\newblock In \emph{International Conference on Machine Learning}, pages
  2144--2155. PMLR, 2021.

\bibitem[Cousins(2021)]{cousins2021axiomatic}
Cyrus Cousins.
\newblock An axiomatic theory of provably-fair welfare-centric machine
  learning.
\newblock \emph{Advances in Neural Information Processing Systems},
  34:\penalty0 16610--16621, 2021.

\bibitem[Dai et~al.(2024)Dai, Xu, Qi, and Jordan]{dai2024incentive}
Xiaowu Dai, Wenlu Xu, Yuan Qi, and Michael Jordan.
\newblock Incentive-aware recommender systems in two-sided markets.
\newblock \emph{ACM Transactions on Recommender Systems}, 2\penalty0
  (4):\penalty0 1--38, 2024.

\bibitem[D'Amour et~al.(2022)D'Amour, Heller, Moldovan, Adlam, Alipanahi,
  Beutel, Chen, Deaton, Eisenstein, Hoffman, et~al.]{d2022underspecification}
Alexander D'Amour, Katherine Heller, Dan Moldovan, Ben Adlam, Babak Alipanahi,
  Alex Beutel, Christina Chen, Jonathan Deaton, Jacob Eisenstein, Matthew~D
  Hoffman, et~al.
\newblock Underspecification presents challenges for credibility in modern
  machine learning.
\newblock \emph{Journal of Machine Learning Research}, 23\penalty0
  (226):\penalty0 1--61, 2022.

\bibitem[Drusvyatskiy and Xiao(2023)]{drusvyatskiy2023stochastic}
Dmitriy Drusvyatskiy and Lin Xiao.
\newblock Stochastic optimization with decision-dependent distributions.
\newblock \emph{Mathematics of Operations Research}, 48\penalty0 (2):\penalty0
  954--998, 2023.

\bibitem[Dwork et~al.(2012)Dwork, Hardt, Pitassi, Reingold, and
  Zemel]{dwork2012fairness}
Cynthia Dwork, Moritz Hardt, Toniann Pitassi, Omer Reingold, and Richard Zemel.
\newblock Fairness through awareness.
\newblock In \emph{Proceedings of the 3rd innovations in theoretical computer
  science conference}, pages 214--226, 2012.

\bibitem[Edgeworth(1881)]{edgeworth1881psychics}
Francis~Ysidro Edgeworth.
\newblock \emph{Mathematical Psychics: An Essay on the Application of
  Mathematics to the Moral Sciences}.
\newblock Kegan Paul, Trench \& Co., London, 1881.

\bibitem[Einav and Rosenfeld(2025)]{einav2025a}
Ohad Einav and Nir Rosenfeld.
\newblock A market for accuracy: Classification under competition.
\newblock In \emph{Forty-second International Conference on Machine Learning},
  2025.

\bibitem[Elkan(2001)]{elkan2001foundations}
Charles Elkan.
\newblock The foundations of cost-sensitive learning.
\newblock In \emph{International joint conference on artificial intelligence},
  volume~17, pages 973--978. Lawrence Erlbaum Associates Ltd, 2001.

\bibitem[Feng et~al.(2024)Feng, Sorensen, Liu, Fisher, Park, Choi, and
  Tsvetkov]{feng2024modular}
Shangbin Feng, Taylor Sorensen, Yuhan Liu, Jillian Fisher, Chan~Young Park,
  Yejin Choi, and Yulia Tsvetkov.
\newblock Modular pluralism: Pluralistic alignment via multi-{LLM}
  collaboration.
\newblock In \emph{Proceedings of the 2024 Conference on Empirical Methods in
  Natural Language Processing}, pages 4151--4171, 2024.

\bibitem[Fish et~al.(2024)Fish, G\"{o}lz, Parkes, Procaccia, Rusak, Shapira,
  and W\"{u}thrich]{fish2024generative}
Sara Fish, Paul G\"{o}lz, David~C. Parkes, Ariel~D. Procaccia, Gili Rusak, Itai
  Shapira, and Manuel W\"{u}thrich.
\newblock Generative social choice.
\newblock In \emph{Proceedings of the 25th ACM Conference on Economics and
  Computation}, EC '24, page 985, New York, NY, USA, 2024. Association for
  Computing Machinery.
\newblock ISBN 9798400707049.

\bibitem[Ge et~al.(2024)Ge, Halpern, Micha, Procaccia, Shapira, Vorobeychik,
  and Wu]{ge2024axioms}
Luise Ge, Daniel Halpern, Evi Micha, Ariel~D. Procaccia, Itai Shapira, Yevgeniy
  Vorobeychik, and Junlin Wu.
\newblock Axioms for {AI} alignment from human feedback.
\newblock In \emph{The Thirty-eighth Annual Conference on Neural Information
  Processing Systems}, 2024.
\newblock URL \url{https://openreview.net/forum?id=cmBjkpRuvw}.

\bibitem[Ghalme et~al.(2021)Ghalme, Nair, Eilat, Talgam-Cohen, and
  Rosenfeld]{ghalme2021strategic}
Ganesh Ghalme, Vineet Nair, Itay Eilat, Inbal Talgam-Cohen, and Nir Rosenfeld.
\newblock Strategic classification in the dark.
\newblock In \emph{International Conference on Machine Learning}, pages
  3672--3681. PMLR, 2021.

\bibitem[Hardt et~al.(2016)Hardt, Megiddo, Papadimitriou, and
  Wootters]{hardt2016strategic}
Moritz Hardt, Nimrod Megiddo, Christos Papadimitriou, and Mary Wootters.
\newblock Strategic classification.
\newblock In \emph{Proceedings of the 2016 ACM conference on innovations in
  theoretical computer science}, pages 111--122, 2016.

\bibitem[Haupt et~al.(2023)Haupt, Hadfield-Menell, and
  Podimata]{haupt2023recommending}
Andreas Haupt, Dylan Hadfield-Menell, and Chara Podimata.
\newblock Recommending to strategic users.
\newblock \emph{arXiv preprint arXiv:2302.06559}, 2023.

\bibitem[Heidari et~al.(2018)Heidari, Ferrari, Gummadi, and
  Krause]{heidari2018fairness}
Hoda Heidari, Claudio Ferrari, Krishna Gummadi, and Andreas Krause.
\newblock Fairness behind a veil of ignorance: A welfare analysis for automated
  decision making.
\newblock \emph{Advances in neural information processing systems}, 31, 2018.

\bibitem[Hicks(1939)]{hicks1939value}
John~R. Hicks.
\newblock \emph{Value and Capital: An Inquiry into Some Foundations of Economic
  Theory}.
\newblock Clarendon Press, Oxford, 1939.

\bibitem[Horowitz and Rosenfeld(2023)]{horowitz2023causal}
Guy Horowitz and Nir Rosenfeld.
\newblock Causal strategic classification: A tale of two shifts.
\newblock In \emph{International Conference on Machine Learning}, pages
  13233--13253. PMLR, 2023.

\bibitem[Horowitz et~al.(2024)Horowitz, Sommer, Koren, and
  Rosenfeld]{horowitz2024classification}
Guy Horowitz, Yonatan Sommer, Moran Koren, and Nir Rosenfeld.
\newblock Classification under strategic self-selection.
\newblock In Ruslan Salakhutdinov, Zico Kolter, Katherine Heller, Adrian
  Weller, Nuria Oliver, Jonathan Scarlett, and Felix Berkenkamp, editors,
  \emph{Proceedings of the 41st International Conference on Machine Learning},
  volume 235 of \emph{Proceedings of Machine Learning Research}, pages
  18833--18858. PMLR, 21--27 Jul 2024.

\bibitem[Hossain et~al.(2025)Hossain, Micha, Chen, and
  Procaccia]{hossain2025strategic}
Safwan Hossain, Evi Micha, Yiling Chen, and Ariel~D. Procaccia.
\newblock Strategic classification with externalities.
\newblock In \emph{The Thirteenth International Conference on Learning
  Representations}, 2025.

\bibitem[Hsu and Calmon(2022)]{hsu2022rashomon}
Hsiang Hsu and Flavio Calmon.
\newblock Rashomon capacity: A metric for predictive multiplicity in
  classification.
\newblock \emph{Advances in Neural Information Processing Systems},
  35:\penalty0 28988--29000, 2022.

\bibitem[Hu and Chen(2020)]{hu2020fair}
Lily Hu and Yiling Chen.
\newblock Fair classification and social welfare.
\newblock In \emph{Proceedings of the 2020 conference on fairness,
  accountability, and transparency}, pages 535--545, 2020.

\bibitem[Jagadeesan et~al.(2023{\natexlab{a}})Jagadeesan, Jordan, and
  Haghtalab]{jagadeesan2023competition}
Meena Jagadeesan, Michael~I Jordan, and Nika Haghtalab.
\newblock Competition, alignment, and equilibria in digital marketplaces.
\newblock In \emph{Proceedings of the AAAI Conference on Artificial
  Intelligence}, volume~37, pages 5689--5696, 2023{\natexlab{a}}.

\bibitem[Jagadeesan et~al.(2023{\natexlab{b}})Jagadeesan, Wei, Wang, Jordan,
  and Steinhardt]{jagadeesan2023learning}
Meena Jagadeesan, Alexander Wei, Yixin Wang, Michael~I Jordan, and Jacob
  Steinhardt.
\newblock Learning equilibria in matching markets with bandit feedback.
\newblock \emph{Journal of the ACM}, 70\penalty0 (3):\penalty0 1--46,
  2023{\natexlab{b}}.

\bibitem[Johansson(1991)]{johansson1991introduction}
Per-Olov Johansson.
\newblock \emph{An introduction to modern welfare economics}.
\newblock Cambridge University Press, 1991.

\bibitem[Jordan(2026)]{jordan2026a}
Michael~I. Jordan.
\newblock A collectivist, economic perspective on ai.
\newblock \emph{Communications of the ACM}, 2026.
\newblock To appear.

\bibitem[Kasy and Abebe(2021)]{kasy2021fairness}
Maximilian Kasy and Rediet Abebe.
\newblock Fairness, equality, and power in algorithmic decision-making.
\newblock In \emph{Proceedings of the 2021 ACM conference on fairness,
  accountability, and transparency}, pages 576--586, 2021.

\bibitem[Keinan and Ben-Porat(2026)]{keinan2026strategic}
Gur Keinan and Omer Ben-Porat.
\newblock Strategic content creation with genai: To share or not to share?
\newblock In \emph{Proceedings of the ACM Web Conference 2026}, pages 169--180,
  2026.

\bibitem[Kleinberg and Raghavan(2020)]{kleinberg2020classifiers}
Jon Kleinberg and Manish Raghavan.
\newblock How do classifiers induce agents to invest effort strategically?
\newblock \emph{ACM Transactions on Economics and Computation (TEAC)},
  8\penalty0 (4):\penalty0 1--23, 2020.

\bibitem[Kleinberg et~al.(2015)Kleinberg, Ludwig, Mullainathan, and
  Obermeyer]{kleinberg2015prediction}
Jon Kleinberg, Jens Ludwig, Sendhil Mullainathan, and Ziad Obermeyer.
\newblock Prediction policy problems.
\newblock \emph{American Economic Review}, 105\penalty0 (5):\penalty0 491--495,
  2015.

\bibitem[Lapin et~al.(2015)Lapin, Hein, and Schiele]{lapin2015top}
Maksim Lapin, Matthias Hein, and Bernt Schiele.
\newblock Top-k multiclass svm.
\newblock \emph{Advances in neural information processing systems}, 28, 2015.

\bibitem[Levanon and Rosenfeld(2021)]{levanon2021strategic}
Sagi Levanon and Nir Rosenfeld.
\newblock Strategic classification made practical.
\newblock In \emph{International Conference on Machine Learning}, pages
  6243--6253. PMLR, 2021.

\bibitem[Levanon and Rosenfeld(2022)]{levanon2022generalized}
Sagi Levanon and Nir Rosenfeld.
\newblock Generalized strategic classification and the case of aligned
  incentives.
\newblock In \emph{International Conference on Machine Learning}, pages
  12593--12618. PMLR, 2022.

\bibitem[Liu(1998)]{liu1998stackelberg}
Baoding Liu.
\newblock Stackelberg-nash equilibrium for multilevel programming with multiple
  followers using genetic algorithms.
\newblock \emph{Computers \& Mathematics with Applications}, 36\penalty0
  (7):\penalty0 79--89, 1998.

\bibitem[Liu et~al.(2018)Liu, Dean, Rolf, Simchowitz, and
  Hardt]{liu2018delayed}
Lydia~T Liu, Sarah Dean, Esther Rolf, Max Simchowitz, and Moritz Hardt.
\newblock Delayed impact of fair machine learning.
\newblock In \emph{International Conference on Machine Learning}, pages
  3150--3158. PMLR, 2018.

\bibitem[Liu et~al.(2022)Liu, Garg, and Borgs]{liu2022strategic}
Lydia~T Liu, Nikhil Garg, and Christian Borgs.
\newblock Strategic ranking.
\newblock In \emph{International Conference on Artificial Intelligence and
  Statistics}, pages 2489--2518. PMLR, 2022.

\bibitem[Liu et~al.()Liu, Yao, Ton, Zhang, Guo, Cheng, Klochkov, Taufiq, and
  Li]{liutrustworthy}
Yang Liu, Yuanshun Yao, Jean-Francois Ton, Xiaoying Zhang, Ruocheng Guo, Hao
  Cheng, Yegor Klochkov, Muhammad~Faaiz Taufiq, and Hang Li.
\newblock Trustworthy {LLM}s: a survey and guideline for evaluating large
  language models' alignment.
\newblock In \emph{Socially Responsible Language Modelling Research}.

\bibitem[Mandi et~al.(2024)Mandi, Kotary, Berden, Mulamba, Bucarey, Guns, and
  Fioretto]{mandi2024decision}
Jayanta Mandi, James Kotary, Senne Berden, Maxime Mulamba, Victor Bucarey, Tias
  Guns, and Ferdinando Fioretto.
\newblock Decision-focused learning: Foundations, state of the art, benchmark
  and future opportunities.
\newblock \emph{Journal of Artificial Intelligence Research}, 80:\penalty0
  1623--1701, 2024.

\bibitem[Marshall(1890)]{marshall1890principles}
Alfred Marshall.
\newblock \emph{Principles of Economics}.
\newblock Macmillan and Co., London, 1890.

\bibitem[Marx et~al.(2020)Marx, Calmon, and Ustun]{marx2020predictive}
Charles Marx, Flavio Calmon, and Berk Ustun.
\newblock Predictive multiplicity in classification.
\newblock In \emph{International Conference on Machine Learning}, pages
  6765--6774. PMLR, 2020.

\bibitem[Mendler-D{\"u}nner et~al.(2024)Mendler-D{\"u}nner, Carovano, and
  Hardt]{mendler-dunner2024an}
Celestine Mendler-D{\"u}nner, Gabriele Carovano, and Moritz Hardt.
\newblock An engine not a camera: Measuring performative power of online
  search.
\newblock In \emph{The Thirty-eighth Annual Conference on Neural Information
  Processing Systems}, 2024.
\newblock URL \url{https://openreview.net/forum?id=hQfcrTBHeD}.

\bibitem[Miller et~al.(2020)Miller, Milli, and Hardt]{miller2020strategic}
John Miller, Smitha Milli, and Moritz Hardt.
\newblock Strategic classification is causal modeling in disguise.
\newblock In \emph{International Conference on Machine Learning}, pages
  6917--6926. PMLR, 2020.

\bibitem[Milli et~al.(2019)Milli, Miller, Dragan, and Hardt]{milli2019social}
Smitha Milli, John Miller, Anca~D Dragan, and Moritz Hardt.
\newblock The social cost of strategic classification.
\newblock In \emph{Proceedings of the Conference on Fairness, Accountability,
  and Transparency}, pages 230--239, 2019.

\bibitem[Mitzenmacher and Vassilvitskii(2022)]{mitzenmacher2022algorithms}
Michael Mitzenmacher and Sergei Vassilvitskii.
\newblock Algorithms with predictions.
\newblock \emph{Communications of the ACM}, 65\penalty0 (7):\penalty0 33--35,
  2022.

\bibitem[Mujica et~al.(2022)Mujica, Crowell, Villano, and
  Uddin]{mujica2022addiction}
Alejandro~L Mujica, Charles~R Crowell, Michael~A Villano, and Khutb~M Uddin.
\newblock Addiction by design: Some dimensions and challenges of excessive
  social media use.
\newblock \emph{Medical Research Archives}, 10\penalty0 (2):\penalty0 1--29,
  2022.

\bibitem[Myerson(1981)]{myerson1981optimal}
Roger~B Myerson.
\newblock Optimal auction design.
\newblock \emph{Mathematics of operations research}, 6\penalty0 (1):\penalty0
  58--73, 1981.

\bibitem[Nahum et~al.(2024)Nahum, Noti, Parkes, and
  Rosenfeld]{nahum2024decongestion}
Omer Nahum, Gali Noti, David~C. Parkes, and Nir Rosenfeld.
\newblock Decongestion by representation: Learning to improve economic welfare
  in marketplaces.
\newblock In \emph{The Twelfth International Conference on Learning
  Representations}, 2024.
\newblock URL \url{https://openreview.net/forum?id=coIaBY8EVF}.

\bibitem[Naseem et~al.(2026)Naseem, Chakraborty, Chang, Dras, Nakov, Peng, and
  Poria]{naseem2026llm}
Usman Naseem, Tanmoy Chakraborty, Kai-Wei Chang, Mark Dras, Preslav Nakov,
  Nanyun Peng, and Soujanya Poria.
\newblock {LLM} alignment should go beyond harmlessness--helpfulness and
  incorporate human agency.
\newblock \emph{Cognitive Computation}, 18\penalty0 (1):\penalty0 26, 2026.

\bibitem[Navon et~al.(2021)Navon, Shamsian, Fetaya, and
  Chechik]{navon2021learning}
Aviv Navon, Aviv Shamsian, Ethan Fetaya, and Gal Chechik.
\newblock Learning the pareto front with hypernetworks.
\newblock In \emph{International Conference on Learning Representations}, 2021.

\bibitem[Ouyang et~al.(2022)Ouyang, Wu, Jiang, Almeida, Wainwright, Mishkin,
  Zhang, Agarwal, Slama, Ray, et~al.]{ouyang2022training}
Long Ouyang, Jeffrey Wu, Xu~Jiang, Diogo Almeida, Carroll Wainwright, Pamela
  Mishkin, Chong Zhang, Sandhini Agarwal, Katarina Slama, Alex Ray, et~al.
\newblock Training language models to follow instructions with human feedback.
\newblock \emph{Advances in neural information processing systems},
  35:\penalty0 27730--27744, 2022.

\bibitem[Pardeshi et~al.(2024)Pardeshi, Shapira, Procaccia, and
  Singh]{pardeshi2024learning}
Kanad~Shrikar Pardeshi, Itai Shapira, Ariel~D Procaccia, and Aarti Singh.
\newblock Learning social welfare functions.
\newblock 2024.

\bibitem[Pareto(1906)]{pareto1906manual}
Vilfredo Pareto.
\newblock \emph{Manual of Political Economy}.
\newblock Macmillan, London, 1906.

\bibitem[Perdomo et~al.(2020)Perdomo, Zrnic, Mendler-D{\"u}nner, and
  Hardt]{perdomo2020performative}
Juan Perdomo, Tijana Zrnic, Celestine Mendler-D{\"u}nner, and Moritz Hardt.
\newblock Performative prediction.
\newblock In \emph{International Conference on Machine Learning}, pages
  7599--7609. PMLR, 2020.

\bibitem[Perdomo(2024)]{perdomo2024the}
Juan~Carlos Perdomo.
\newblock The relative value of prediction in algorithmic decision making.
\newblock In \emph{Proceedings of the 41st International Conference on Machine
  Learning (ICML)}, 2024.
\newblock URL \url{https://openreview.net/forum?id=oaACFfNbXl}.

\bibitem[Petersen et~al.(2022)Petersen, Kuehne, Borgelt, and
  Deussen]{petersen2022differentiable}
Felix Petersen, Hilde Kuehne, Christian Borgelt, and Oliver Deussen.
\newblock Differentiable top-k classification learning.
\newblock In \emph{International Conference on Machine Learning}, pages
  17656--17668. PMLR, 2022.

\bibitem[Pigou(1920)]{pigou1920welfare}
Arthur~Cecil Pigou.
\newblock \emph{The Economics of Welfare}.
\newblock Macmillan, London, 1920.

\bibitem[Rafailov et~al.(2023)Rafailov, Sharma, Mitchell, Manning, Ermon, and
  Finn]{rafailov2023direct}
Rafael Rafailov, Archit Sharma, Eric Mitchell, Christopher~D Manning, Stefano
  Ermon, and Chelsea Finn.
\newblock Direct preference optimization: Your language model is secretly a
  reward model.
\newblock \emph{Advances in Neural Information Processing Systems},
  36:\penalty0 53728--53741, 2023.

\bibitem[Rodolfa et~al.(2020)Rodolfa, Salomon, Haynes, Mendieta, Larson, and
  Ghani]{rodolfa2020case}
Kit~T Rodolfa, Erika Salomon, Lauren Haynes, Iv{\'a}n~Higuera Mendieta, Jamie
  Larson, and Rayid Ghani.
\newblock Case study: predictive fairness to reduce misdemeanor recidivism
  through social service interventions.
\newblock In \emph{Proceedings of the 2020 Conference on Fairness,
  Accountability, and Transparency}, pages 142--153, 2020.

\bibitem[Rolf et~al.(2020)Rolf, Simchowitz, Dean, Liu, Bjorkegren, Hardt, and
  Blumenstock]{rolf2020balancing}
Esther Rolf, Max Simchowitz, Sarah Dean, Lydia~T Liu, Daniel Bjorkegren, Moritz
  Hardt, and Joshua Blumenstock.
\newblock Balancing competing objectives with noisy data: Score-based
  classifiers for welfare-aware machine learning.
\newblock In \emph{International Conference on Machine Learning}, pages
  8158--8168. PMLR, 2020.

\bibitem[Rudin(2019)]{rudin2019stop}
Cynthia Rudin.
\newblock Stop explaining black box machine learning models for high stakes
  decisions and use interpretable models instead.
\newblock \emph{Nature machine intelligence}, 1\penalty0 (5):\penalty0
  206--215, 2019.

\bibitem[Saig et~al.(2023)Saig, Talgam-Cohen, and Rosenfeld]{saig2023delegated}
Eden Saig, Inbal Talgam-Cohen, and Nir Rosenfeld.
\newblock Delegated classification.
\newblock \emph{Advances in Neural Information Processing Systems},
  36:\penalty0 13200--13236, 2023.

\bibitem[Sen(1970)]{sen1970collective}
Amartya Sen.
\newblock \emph{Collective Choice and Social Welfare}.
\newblock Harvard University Press, Cambridge, MA, 1970.

\bibitem[Sen(1999)]{sen1999commodities}
Amartya Sen.
\newblock Commodities and capabilities.
\newblock \emph{OUP Catalogue}, 1999.

\bibitem[Shapley and Shubik(1971)]{shapley1971assignment}
Lloyd~S Shapley and Martin Shubik.
\newblock The assignment game i: The core.
\newblock \emph{International Journal of game theory}, 1\penalty0 (1):\penalty0
  111--130, 1971.

\bibitem[Shavit et~al.(2020)Shavit, Edelman, and Axelrod]{shavit2020causal}
Yonadav Shavit, Benjamin Edelman, and Brian Axelrod.
\newblock Causal strategic linear regression.
\newblock In \emph{International Conference on Machine Learning}, pages
  8676--8686. PMLR, 2020.

\bibitem[Sherali(1984)]{sherali1984multiple}
Hanif~D Sherali.
\newblock A multiple leader stackelberg model and analysis.
\newblock \emph{Operations Research}, 32\penalty0 (2):\penalty0 390--404, 1984.

\bibitem[Shirali et~al.(2024)Shirali, Abebe, and Hardt]{shirali24allocation}
Ali Shirali, Rediet* Abebe, and Moritz* Hardt.
\newblock Allocation requires prediction only if inequality is low.
\newblock In \emph{Proceedings of the 41st International Conference on Machine
  Learning (ICML)}. Proceedings of Machine Learning Research (PMLR), July 2024.
\newblock URL \url{https://proceedings.mlr.press/v235/shirali24a.html}.
\newblock *equal contribution.

\bibitem[Sommer et~al.(2025)Sommer, Hikri, Amit, and
  Rosenfeld]{sommer2025learning}
Yonatan Sommer, Ivri Hikri, Lotan Amit, and Nir Rosenfeld.
\newblock Learning classifiers that induce markets.
\newblock In \emph{Forty-second International Conference on Machine Learning},
  2025.

\bibitem[Sorensen et~al.()Sorensen, Moore, Fisher, Gordon, Mireshghallah,
  Rytting, Ye, Jiang, Lu, Dziri, et~al.]{sorensenposition}
Taylor Sorensen, Jared Moore, Jillian Fisher, Mitchell~L Gordon, Niloofar
  Mireshghallah, Christopher~Michael Rytting, Andre Ye, Liwei Jiang, Ximing Lu,
  Nouha Dziri, et~al.
\newblock Position: A roadmap to pluralistic alignment.
\newblock In \emph{Forty-first International Conference on Machine Learning}.

\bibitem[Stiglitz(2012)]{stiglitz2012inequality}
Joseph~E. Stiglitz.
\newblock \emph{The Price of Inequality: How Today's Divided Society Endangers
  Our Future}.
\newblock W.W. Norton \& Company, New York, 2012.

\bibitem[Wager and Athey(2018)]{wager2018estimation}
Stefan Wager and Susan Athey.
\newblock Estimation and inference of heterogeneous treatment effects using
  random forests.
\newblock \emph{Journal of the American Statistical Association}, 113\penalty0
  (523):\penalty0 1228--1242, 2018.

\bibitem[Wang et~al.(2024)Wang, Bates, Aronow, and
  Jordan]{wang2024counterfactual}
Serena Wang, Stephen Bates, P~Aronow, and Michael Jordan.
\newblock On counterfactual metrics for social welfare: Incentives, ranking,
  and information asymmetry.
\newblock In \emph{International Conference on Artificial Intelligence and
  Statistics}, pages 1522--1530. PMLR, 2024.

\bibitem[Wei et~al.(2026)Wei, Shi, and Zhang]{wei2026market}
Xiukun Wei, Min Shi, and Xueru Zhang.
\newblock Market games for generative models: Equilibria, welfare, and
  strategic entry.
\newblock In \emph{The Fourteenth International Conference on Learning
  Representations}, 2026.

\bibitem[Yao et~al.(2023)Yao, Li, Nekipelov, Wang, and Xu]{yao2023bad}
Fan Yao, Chuanhao Li, Denis Nekipelov, Hongning Wang, and Haifeng Xu.
\newblock How bad is top-$ k $ recommendation under competing content creators?
\newblock In \emph{International Conference on Machine Learning}, pages
  39674--39701. PMLR, 2023.

\bibitem[Yao et~al.(2024)Yao, Liao, Wu, Li, Zhu, Yang, Liu, Wang, Xu, and
  Wang]{yao2024user}
Fan Yao, Yiming Liao, Mingzhe Wu, Chuanhao Li, Yan Zhu, James Yang, Jingzhou
  Liu, Qifan Wang, Haifeng Xu, and Hongning Wang.
\newblock User welfare optimization in recommender systems with competing
  content creators.
\newblock In \emph{Proceedings of the 30th ACM SIGKDD Conference on Knowledge
  Discovery and Data Mining}, pages 3874--3885, 2024.

\end{thebibliography}

\newpage
\appendix
\section{Broader impact}

Any academic discipline with the capacity to influence social outcomes 
should do so with much care and deliberation. 
For machine learning,
this position paper proposes that social welfare serve as a core guiding principle for the design of socially-beneficial learning algorithms.
In principle, and by construction, the impact of our proposed framework on society aims to be positive. But our paper also points to the inherent difficulties of identifying---and formalizing---what `good for society' is.
Given that welfare economics has grappled with this very challenge for decades, 
its perspective offers valuable guidance for the learning community towards this daunting task.
Nonetheless, while the ideas underlying the social welfare framework are conceptually appealing,
the path to improving social welfare in practice is likely to surface many additional challenges---%
some expected, others unforeseen.

For example, we may specify incorrect social welfare functions;
or specify them correctly, but fail to optimize them effectively;
overcome optimization challenges---only to find that 
our assumptions are flawed, that theory diverges from practice,
or that important factors were overlooked.
Adapting existing machine learning methodology
to support welfare notions such as resources, allocations, agency, and social policy is unlikely to be straightforward.
However, as we argue throughout, we believe that building on existing machinery is more viable and practical than starting from scratch.
The necessity of confronting the challenges that welfare presents can also
be taken as an opportunity---to drive machine learning research towards more informed, transparent, responsible, and socially aware practice.


\paragraph{Who drives welfare?}
In the late 1940s, the field of economics found itself in a position of significant social influence, as newly developed theoretical ideas and practical tools began to exert unprecedented impact on society.
Eighty years later,
machine learning appears to be following a similar trajectory.
It is therefore only natural to try and learn from these earlier precedents---both successful and less so.
One lesson is that the wheel need not be (re)invented, but rather adapted.
While machine learning now exerts perhaps comparable influence---despite not being designed to do so---welfare economics was explicitly developed to support policymakers.
This is one grounds for our argument of drawing on ideas from welfare economics.

A second important lesson learned is that economic objectives,
welfare included,
can rarely be forced upon a system.
Economics views welfare as an emergent outcome arising from the decentralized interactions of many agents,
particularly in idealized settings such as markets with perfect competition;
this is the `invisible hand' in essence.
But in realistic setting, welfare needs guidance to materialize,
whether through legislation, incentive alignment, or even minor nudges.
As theoreticians, our goal should be to understand how systems can be steered towards beneficial outcomes, and what market failures can occur absent such direction.
As system designers, we should carefully avoid coercing our own beliefs,
and instead enable policymakers to implement their policies \emph{through} the tools we build while maintaining transparency and accountability.
As practitioners, we should strive to use machine learning to create real economic value for society at large.


\paragraph{Who should care?} We believe the following target audiences,
spanning different levels of familiarity with economics,
may all find our position helpful.
The first, and perhaps foremost, audience are general ML researchers and practitioners working on problems that have downstream social impact---%
but who do not explicitly account for this in their work. 
Clear example applications include recommendation systems, social networks, media platforms, and e-commerce systems. However, we anticipate that over time, more and more subfields of ML will find themselves in a position where their work gains increasing social impact; consider text and image generation as such recent examples. 
It is better to plan for this ahead than facing serious welfare issues later.

For this group,
our goal is not only to raise awareness to welfare considerations,
but also to demonstrate that incorporating them is within their reach.
This is why we build our examples around supervised learning:
it is fundamental to ML, widely familiar, and ubiquitous in practice,
yet raises elementary welfare questions at different levels of economic abstraction.
Our vision of, for example, benchmarks in which participants compete over who generates greater social value can materialize only if people who already have the necessary mindset, skills, expertise to build such environments contribute to these efforts.
Ultimately, we believe that welfare will prove effective for machine learning only if it is embraced and grows organically within the broad community.

The second (smaller) audience consists of researchers in sub-fields that already draw some connections to welfare or certain aspects of it,
such as fairness, strategic learning, performative prediction, and, increasingly, generative model alignment.
For this group, we aim to provide a broader, unifying framework to the questions they are already pursuing, 
bridging these fields together under a common organizing principle
to illuminate their similarities and differences.
Welfare can also offer a complementary perspective on topics that may have come to dominate the discourse.
For example, and as we argue, fairness research largely overlooks agency, 
strategic classification rarely considers resources,
and performative prediction prioritizes accuracy.
By revealing such welfare ``blind spots'', we hope to encourage the exploration of new and important research directions.

The final (and even smaller) audience consists of researchers already working at the intersection of machine learning and economics. 
This group is likely well-versed in notions of social welfare
and its underlying philosophy.
Our goal here is to enable better integration of ideas from economics within machine learning research and practice.
To have impact, we believe that work at this intersection must retain
the core elements of learning problems.
The view of welfare we advance is therefore carefully crafted to foster effective communication of ideas between machine learning and economics in \emph{both} directions.
Doing so can help establish economic considerations 
as central to the field, while also
lowering the barrier of entry for interested researchers and practitioners
with limited background in economics. 







\section{Social welfare functions} \label{appx:SWFs}

The literature on social welfare functions is rich and diverse. 
Here we present a basic and simplified subset of alternative which can help shed further light on our proposed framework. Interested readers are referred to \citep{adler2019measuring} for a more thorough discussion of possible alternatives.
For simplicity we present welfare as a function only of utilities,
i.e., without planner-determined example weights $w(x)$.
We note however, that weights can more generally depend also on the utility $u(x)$ (when it is known).
This allows expressing welfare using broader functional forms
and relative measures of inequality (e.g., Gini, Theil). 
For clarity, we present empirical welfare definitions, rather than their expected analog.
\squeeze


As we note, the most basic social welfare function is \emph{utilitarian welfare}:
\[
\welf = \frac{1}{n} \sum_{i=1}^n  u_i
\]
This assigns equal weights to all individuals, and aims to maximize average user benefits.

To prioritize equity, \emph{egalitarian welfare functions} typically penalize 
for dispersion across utility, for example:
\[
\welf = \frac{1}{n} \sum_{i=1}^n  u_i - \lambda \cdot \mathtt{var}(u_1,\dots,u_n)
\]
where $\lambda$ explicitly trades off (utilitarian) efficiency with equity.

An alternative is \emph{Nash social welfare},
which replaces the arithmetic mean with the geometric mean:
\[
\welf = \Big( \prod_{i =1}^n  u_i \Big)^{1/n}
\]
This enables a particular trade-off of efficiency and equity
through control of the dispersion of utility across users.
For example, when some users have significantly higher values than other,
the geometric mean ``rescales'' their importance and reduces their influence.


Some social welfare functions follow from the Rawlsian theory of justice,
whose main concern is the utility of the worst-off individual in society:
\[
\welf = \min\{u_1, u_2, \ldots, u_n\}
\] 
From a welfare perspective, this sets the goal of maximizing the minimal utility across users.

Note the above permits flexibility in setting preferences over outcomes for others.
Hence, and due to it's extremist nature, some social welfare functions in this category
permit relaxing its strict maxmin nature if this provides significant benefits across the population. One example are \emph{leximin welfare functions}, such as:
\[
\welf = \frac{1}{n} \sum_{i =1}^n \alpha^{i}  u_{(i)}
\]
where the $u_{(i)}$ notation denotes users indexed by ranking their utility
(i.e., $u_{(i)}$ is the $i_{th}$ lowest utility),
and $\alpha \in (0,1)$ is a design parameter.
This simple weighing scheme is a simple and straightforward way to implement
the Rawlsian perspective.

\section{Further prospects and challenges} \label{appx:beyond}
Social welfare offers many benefits, but also poses significant challenges.
Here we detail prospects and challenges beyond those discussed in the paper.


\subsection{Regulation: Social planner, revisited} \label{sec:regulation}
So far we have used the notion of `social planner' mostly as
a useful construct for defining the social preferences that should guide learning.
This has aided us in discussing how to learn in a ways that accommodates certain given preferences, and in designing learning algorithms that support flexible policies.
But there are additional roles for the social planner to play,
requiring different interfaces with the learning process.
For example:
\squeeze
\begin{itemize}[leftmargin=1em,topsep=0em,itemsep=0.3em]

\item \textbf{Monitoring.}
One defining characteristic of data-driven systems
is that they hold a distinct informational advantage over their users.
In economics, information asymmetry is a well-known source of hazzard for markets, referred to as \emph{adverse selection} \cite{akerlof1978market}.
Monitoring can help to mitigate this by identifying what information gives the system an unfair advantage, and then channeling this information to users.
This provides an operational motivation for transparency,
with the definitive purpose of improving welfare.



\priority{%
\item \textbf{Auditing.}
- random cost
- risk aversion of users (vs system that plays for expectation)
}

\item \textbf{Subsidizing.}
When resources are limited, their allocation often depends on the wealth or assets of individuals.
This presents an opportunity to intervene by injecting funds into the system in a way that balances outcomes, e.g., via targeted subsidies.
Subsidies can either be given in response to a learned classifier,
or determined by policy prior to learning.
Since outcomes in our case depend on predictions,
this presents a unique opportunity for a social planner to intervene by directly providing users with useful `features',
thus bypassing the need for a monetary mechanism.
An interesting question is then how to design learning objectives that operate under such subsidies.
\extended{\tocitec{scgk}}

\priority{%
\item \textbf{Stress-testing.}
- counterfactual (ref stress-test paper)
}

\end{itemize}

Further distinct yet important roles include:
setting standards, licensing, auditing,
mandating disclosure and transparency, and stress-testing.

\subsection{Inverse tasks: policy preference elicitation}
A key point we have made throughout is that 
any prediction policy advances \emph{some} social preferences.
Until now, our focus was on learning to maximize welfare under a specified social welfare function.
But we can also ask the inverse question:
given a policy---what social welfare function is it optimizing?
This can be useful for revealing the preferences of general,
unspecified policies,
and for evaluating whether a policy that has been specified truly achieves its goals.
One way to approach this would be to use tools from econometrics
(e.g. as in \citep{bjorkegren2022machinelearningpoliciesvalue,pardeshi2024learning}).
But is also interesting to ask whether ideas from inverse problems in machine learning can be adopted.



\subsection{Evaluation}
A main driving force of modern machine learning is the ability to evaluate the performance of different methods effectively and consistently under a wide range of settings and conditions.
Unfortunately, such privileges are not possible when 
learning in social settings.
The challenge of evaluation is shared by all fields at the intersection of learning, economics, and human behavior,
but welfare-maximizing machine learning presents its own unique difficulties.
For example, even the basic task of measuring or estimating social welfare is far from trivial. 
One immediate challenges is the difficulty (and restrictions) of obtaining and working with representative and informative user data.
Another is that, as we have stressed, welfare depends on outcomes, which are the product of decisions due to policy,
and so evaluation requires either control of the policy or a way to estimate counterfactuals.
Other than a few noteworthy exceptions \citep[e.g.,][]{bjorkegren2020manipulation,haupt2023recommending,mendler-dunner2024an},
most current works settle for `semi-synthetic' evaluation that uses real data and simulated user behavior.
Simulation is often useful as an intermediary step to drive the field forward (e.g., as in reinforcement learning),
but the ultimate goal should be to enable seamless, realistic, in-the-wild evaluation.




\subsection{Accounting for causal effects.}  \label{appx:causal}

In Sec.~\ref{sec:lens-explicit} we consider policies of a particular type,
called \emph{prediction policies} \cite{kleinberg2015prediction}, in which actions are determined by predictions as $a = \policy_h(x) = \policy(h(x))$.
The key assumption of prediction policies is that knowing the true $y$ suffices for making an optimal decision; hence the motivation for replacing $y$ with the prediction $\yhat=h(x)$.
But in realistic settings, $y$ is rarely the only source of uncertainty.
Most decision settings have a casual aspect,
meaning that actions $a$ can \emph{affect} outcomes $y$.
Hence, there is no `true' $y$ on which we can draw,
and the optimal action $a$ (or `intervention')
must include an understanding of its causal effects on $y$.

When learning in social settings, the issue of causality is further complicated 
by the fact that users have agency, and therefore act.
Because learning determines policy, and policies shape users' choice behavior,
learning comes to have a second-order causal effect on $y$.
The task of causal inference therefore attains two goals:
understand the impact of decisions on outcomes directly,
and indirectly through user choices.

Econometrics offers many tools for contending with questions of 
causal estimation.
The main challenge is to make inference of causal effects on the basis of observed data---which is the common setting considered in learning.
In cases where we can experiment, randomized control trials (or A/B tests)
offer an alternative.
Some of these ideas have been used in the literature on strategic learning
\citep{miller2020strategic,shavit2020causal,horowitz2023causal},
mostly in relation to the interaction between the system and its users.
A prominent point that concerns welfare is that a causal understanding can be used
to \emph{improve} outcomes---if incentives are set correctly
\citep{kleinberg2020classifiers}.
Despite the importance of causal effects,
we believe there is merit in first exploring question of welfare
under policy predictions, as means to set the stage for more advanced questions concerning causality.



\end{document}